\documentclass[letterpaper, 10 pt, journal, twoside]{IEEEtran}  

\IEEEoverridecommandlockouts                              

\usepackage{amsmath} 
\usepackage{amssymb}  
\usepackage{caption}
\usepackage{subcaption}
\usepackage[]{algorithm2e}
\SetKwProg{Fn}{Function}{}{}
\usepackage{pdfpages}

\author{Philippe Morere$^{1}$, Lionel Ott$^{1}$ and Fabio Ramos$^{1,2}$
\thanks{This paper was published in IEEE Robotics and Automation Letters (2019).}
\thanks{$^{*}$Correspondence to {\tt\small philippe.morere@sydney.edu.au}}%
\thanks{$^{1}$The University of Sydney}%
\thanks{$^{2}$NVIDIA}%
}

\title{Learning to Plan Hierarchically from Curriculum}

\begin{document}

\maketitle

\begin{abstract}
We present a framework for learning to plan hierarchically in domains with unknown dynamics.
We enhance planning performance by exploiting problem structure in several ways:
(i) We simplify the search over plans by leveraging knowledge of skill objectives,
(ii) Shorter plans are generated by enforcing aggressively hierarchical planning,
(iii) We learn transition dynamics with sparse local models for better generalisation.
Our framework decomposes transition dynamics into skill effects and success conditions, which allows fast planning by reasoning on effects, while learning conditions from interactions with the world. We propose a simple method for learning new abstract skills, using successful trajectories stemming from completing the goals of a curriculum. Learned skills are then refined to leverage other abstract skills and enhance subsequent planning. We show that both conditions and abstract skills can be learned simultaneously while planning, even in stochastic domains.
Our method is validated in experiments of increasing complexity, with up to $2^{100}$ states, showing superior planning to classic non-hierarchical planners or reinforcement learning methods. Applicability to real-world problems is demonstrated in a simulation-to-real transfer experiment on a robotic manipulator.
\end{abstract}

\section{INTRODUCTION}
\IEEEPARstart{S}{ample} efficiency is of utmost importance when robots need to learn how to act using interactions with their environment. Reinforcement learning (RL) methods are especially affected by this issue, and often require very large amounts of data before learning decent policies. This is unacceptable in many robotics scenarios where gathering data is expensive or time consuming. This problem can largely be addressed by taking advantage of structure in states, actions and environment transitions. However, learning this structure is very challenging.

\begin{figure}
\centering
\includegraphics[width=0.8\columnwidth]{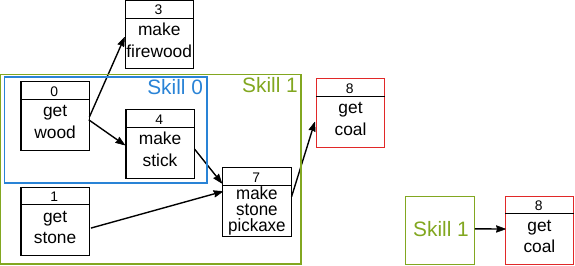}
\caption{Left: learning abstract skills (blue and green rectangles) from a sequence of primitive skills (boxes), after planning for a given goal (red box). Right: resulting abstracted trajectory for goal \emph{get coal}.\label{fig:graph_skill_refine}}
\end{figure}

Many robotic planning problems feature a hierarchical task structure; e.g. graph in Figure \ref{fig:graph_skill_refine}. Taking advantage of problem hierarchy by planning at abstract levels leads to multiple advantages over classic planners or RL techniques which operate at a single level only. Indeed, plans constructed from abstract actions, or \emph{skills}, are typically much shorter as they can reuse previous skills. Hierarchical plans can also be lazy, ie. skills are only decomposed into lower-level skills when needed, allowing plans to be updated with the latest environment information.
Furthermore, planning with high-level skills is often easier, as most of the environment stochasticity is absorbed into lower-level skills. These principles greatly improve sample efficiency and planning times compared to RL and classic planners, as shown in experiments.

When deploying robots in the real world, it is common to use a set of pre-defined policies to perform basic actions instead of using raw torque commands for example. These policies can either be programmed by experts or learned using RL techniques for example, to perform a specific task. Because the effect of these pre-defined policies (or \emph{primitive skills}) is usually well-defined, it can be leveraged by planning algorithms to enhance their efficiency. However, in many cases, the conditions under which primitive skills succeed may be unknown, and must be learned from interactions with the environment. Knowledge of skill effects enables directed planning towards a given goal, and makes inefficient random exploration unnecessary. Our experiments confirm this, showing directed planning is orders of magnitude faster than Monte-Carlo Tree Search~\cite{coulom2006efficient} and RRT~\cite{lavalle1998rapidly}.


We develop a framework for learning to plan hierarchically in domains with unknown dynamics. This framework aims to provide planning algorithms with more structure, and is based on several key insights.
Firstly, we consider the problem of planning with primitive skills of known effects. Skill effects can be exploited by planning algorithms to determine the best available skill for a given goal. Using a \emph{curriculum}, a sequence of goals with increasing complexity, new useful skills are incrementally made available to learn.
Secondly, interesting and challenging problems often feature a hierarchical component which flat planning algorithms struggle to cope with. Our method has a strong emphasis on hierarchical planning, and encourages learned skills to reuse other skills. This results in efficient plans with very few high-level steps, as shown in Figure~\ref{fig:graph_skill_refine}.
Lastly, in structured and sparse state spaces (ie. state dimensions are mostly independent of each other), skill effects, skill conditions and transition dynamics also become sparse. In such environments, transition dynamics or skill conditions can then be learned using very simple local models, allowing scalability to problems with larger state spaces. Using local models for transitions dynamics also improves generalisation to unseen states as most state dimensions become irrelevant to predict specific transitions.

Our contributions are the following:
\begin{itemize}
\item We present a framework suited for hierarchical planning, in which transition dynamics are decomposed into skill effects and conditions. This framework allows for reasoning on skill effects, while learning their conditions from interactions with the world.
\item We propose a simple method for learning new abstract skills, using successful trajectories stemming from completing the goals of a curriculum. Skills are then refined by reasoning on the effects and conditions of previous successful trajectories.
\item We extend the problem to the case of unknown transition dynamics, ie. when skill success conditions are unknown. We present a method for learning conditions from interactions with the real world, and show that conditions can be learned while planning, learning and refining skills, even in stochastic environments.
\item We evaluate our approach performance on simulated problems of growing complexity against established planners and RL methods. We then demonstrate the applicability to real-world problems with a simulation-to-real transfer experiment on a robotic manipulator.
\end{itemize}


\section{RELATED WORK}
\label{sec:related_work}
The idea of planning by reasoning about action effects and conditions has been long studied, and is a core principle behind classic planners like STRIPS~\cite{fikes1971strips}. These planners typically require all action effects and conditions to be specified, and often produce sequential plans, thus lacking the advantages of hierarchical planning. Hierarchical Task Networks (HTN)~\cite{nau1999shop} provide a hierarchical alternative by producing plans given skills and their dependencies.
HTN planners, and their extension to AND/OR graphs~\cite{de1990and}, have been successfully applied to rich and complex problems such as robot soccer~\cite{obst2005using}, multi-agent assembly domains~\cite{knepper2014distributed}, and human-robot collaborative assembly~\cite{johannsmeier2017hierarchical}.
However, these planners require dynamics and hierarchy to be specified.

Learning the skill hierarchy while planning with HTN is proposed in~\cite{nejati2006learning}, but relies on expert demonstrations, and transition dynamics still need to be specified.
More recent work requires a graphical task representation to automatically construct HTMs~\cite{hayes2016autonomously}. These techniques necessitate considerable expert knowledge. By opposition, our work aims to learn both dynamics and hierarchy directly from interactions with the world.

Another approach to plan in environments with unknown dynamics is reinforcement learning (RL).
In model based RL~\cite{sutton1991dyna}, environment dynamics are learned from observed transitions, and the learned model can then be used in conjunction with a planning algorithm such as Monte-Carlo Tree Search~\cite{coulom2006efficient}, DESPOT~\cite{ye2017despot} or RRT \cite{lavalle1998rapidly,garrett2017sample} to find an optimal plan. Although powerful, these methods are not robust against small errors in learned transition dynamics, as these compound when planning over long horizons. By opposition, hierarchical planners typically plan on much shorter horizons, which mitigates this problem.

The idea of hierarchical learning and planning was also studied in RL~\cite{sutton1999between}. In hierarchical RL, policies are composed sub-policies called \emph{options}, which can be learned from interactions with the environment.
Symbolic planning and RL were combined in~\cite{grounds2008combining} where a STRIPS planner shapes the RL agent's reward function, to achieve high-level reasoning and fast low-level reactions.
The HASSLE algorithm~\cite{bakker2004hierarchical} learns sub-skills by identifying clusters of raw input data, but is limited to a small and predefined number of hierarchical levels.
Unlike our work, RL methods learn policies using a fixed reward signal, and thus often can't handle multiple or changing goals.

When dealing with high dimensional state and action spaces, planning and RL performance degrades quickly.
Relational RL~\cite{van2005survey} is a subset of RL concerned with upgrading state and action representations with objects and their relations. These MDP variants can deal with very high dimensional and structured state spaces, by reasoning over objects.
The transition dynamics can be compactly represented using dynamic Bayesian networks or decision trees~\cite{boutilier2000stochastic}.
Our work addresses high dimensional state and action spaces with sparsity assumption, although it could be extended to using relational RL concepts.

Planning with policies or skills instead of actions was investigated in~\cite{konidaris2018skills}, where skills are black-box controllers. The problem is extended to the case of parametrised skills by~\cite{ames2018learning}, which improves the range of available robot behaviours at the expense of requiring planning over both skill and parameters. Lastly, actions are replaced with predefined \emph{algorithms} in~\cite{tavares2018algorithms}, yielding advantages over planning with action directly.

This work combines the advantages of hierarchical planners such as HTN with the ability to learn transition dynamics from interactions with the environment, as achieved by RL algorithms. The presented framework leverages sparse state representations, allowing it to scale to problems with very large state spaces.

\section{PROBLEM DEFINITION}
\label{sec:problem_definition}
We begin by defining the family of problems addressed by our method, and present a general framework for hierarchical planning in which abstract skills and primitive skill conditions can be learned from interactions with the environment.

\subsection{Markov environment}
We consider an agent interacting with a Markov environment. The environment is a tuple $(\mathcal{S}, \mathcal{A}, \mathcal{G}, T)$ composed of states $s \in \mathcal{S}$, skills $a \in \mathcal{A}$, goals $g \in \mathcal{G}$, and a transition function $T$.
States $s$ are vectors representing the current full state of the environment, and the environment's Markov property restricts transitions to a new state $s'$ to only depend on the previous state $s$ and skill $a$, ie. $s' = T(s,a)$. This property greatly simplifies planning, as maintaining the history of previous states is not required. Agents are given a goal $g$ and must find a sequence of skills (called \emph{plan}) to reach $g$, starting from an initial state $s_0$. Solving the planning problem reduces to finding plans $\pi$ that reach \emph{any} feasible goal given to the agent.

\subsection{Framework}
\label{sec:framework}
We extend the Markov environment definition to incorporate skill knowledge at its core. The set of initial skills given to the agent is denoted $\mathcal{A}^0$. Skills in $\mathcal{A}^0$ are called \emph{primitive} skills, as opposition to \emph{learned} skills which are learned from successful trajectories at a later stage.

\subsubsection{Skills}
\label{sec:skill_definition}
A skill $a$ (primitive or learned) is composed of the following elements:
\begin{itemize}
\item An effect $e$ describes the intended state changes resulting from executing the skill. For example, an effect can be a set of state dimensions to be changed and their resulting values.
\item A condition $c$ characterises state requirements for a skill to succeed. When these conditions are fulfilled, the skill can be successfully executed, and effect $e$ is applied to the new state. If these conditions are not met, skill execution fails and the effect $e$ is not applied (although $e$ could still be observed due to stochasticity).
\item A plan or policy $\pi$. In the case of primitive skills, $\pi$ is a predefined policy or a raw action, which can be directly executed in the environment. For learned skills, $\pi$ is a sequence of skills (learned and/or primitive) which when executed achieves the desired effect $e$.
\item A side effect $e^+$ describes additional non-intended state changes resulting from executing $\pi$. Primitive skills often have few or no side effects, whereas learned skills with complicated plans may trigger multiple side effects upon execution.
\end{itemize}

\subsubsection{Transitions}
Environment transitions are decomposed in a similar way to skill effects and conditions.
The transition function $T$ is modelled as the composition of an effect function $f$ and a noise function $g$:
\begin{align}
s' = T(s,a) &= g(f(s, a)),\\
\text{where } f(s, a) &= \begin{cases}
					 	  s \oplus e \text{ if } c \text{ matches } s,\\
					 	  s \text{ otherwise},
                         \end{cases}\\
\text{and } g(s) &= s \oplus \epsilon. \label{eq:stochastic_transition}
\end{align}
Here $e$ and $c$ refer to the effect and condition of skill $a$ respectively, and $s \oplus e$ indicates that effect $e$ is applied to state $s$. Function $g$ reflects potentially stochastic dynamics by applying noise effect $\epsilon$ to the resulting state of $f(s,a)$.
Skill conditions can either be given to the agent, in which case the problem reduces to strict planning, or they may be unknown (which translates to unknown transition dynamics) and need to be learned from interactions with the environment.

\subsubsection{Sparse states}
Keeping scalability to larger problem in mind, we enforce sparse state representations; eg. as binary vectors of features. Sparse state representations make skill effects and conditions compact, as each skill only operates on a few dimensions of the state space. This sparse requirement results in easier condition learning, and allows reasoning and planning even in high dimensional state spaces.
Although this is not the scope of this paper, learning a disentangled state representation could be achieved using auto-encoders \cite{thomas2017independently}.

The presented decomposition of skills and transition dynamics into conditions and effects enables much richer reasoning over action intentions and goals. The method presented in the next section leverages this principle to learn skills, conditions, and plan hierarchically.

\section{METHOD}
\label{sec:method}
We present a method for learning to plan hierarchically\footnote{Python code available at {https://github.com/PhilippeMorere/learning-to-plan-hierarchically}}, following the framework defined in Section~\ref{sec:framework}.
The proposed technique plans backwards from desired goal to starting state, using a collection of skills $\mathcal{A}$. Skills are composed with one another by matching the effects of a skill with the conditions of the next. Plans successfully achieving desired goals are abstracted into new skills, then added to $\mathcal{A}$. New skills are refined to reuse other skills of $\mathcal{A}$, so as to ensure planning results in short and highly hierarchical plans. This refinement process relies on reasoning about skill effects and conditions; in problems with unknown dynamics, skill conditions can be learned from interactions with the environment, using a probabilistic model.

\subsection{Planning with hierarchical skills}
\label{sec:hierarchical_planning}
The presented planner takes advantage of the rich collection of skills $\mathcal{A}$ and their hierarchical nature. Planning backwards, it recursively finds skills that satisfy the conditions of its current goal. Starting from goal $g$, the planner finds the skill $a \in \mathcal{A}$ with effect closest to $g$, sets $a$'s success conditions as its new goal, and plans again. If no skill exactly matches $g$, the returned plan -- composed of the skill whose effect is closest to $g$ -- approaches the vicinity of $g$. This helps subsequent planning (or random skills) to reach $g$.
The hierarchical goal-regression planner is detailed in Algorithm~\ref{alg:hierarchical_planner}.

This framework promotes aggressive hierarchical planning and leads to very fast planning, as the maximum recursion $rec_{max}$ can be low ($rec_{max}=3$ was sufficient in experiments). Also, this hierarchical planner returns lazy plans; plans are never reduced to sequences of primitive skills. Thus executing the first primitive skill of a plan only requires expanding its first element. This property allows plans to be adapted automatically by expanding higher-level skills only when the newest state is available, hence making plans more robust to stochastic transitions and unforeseen side effects.

\subsection{Skill learning}
\label{sec:skill_learning}
Planning performance greatly depends on the quality and diversity of skills in $\mathcal{A}$; augmenting $\mathcal{A}$ using successful plans is essential.
Skill learning only requires a successful trajectory and an intended goal, and thus equally applies to previous successful plans and expert demonstrations. Skills can also be learned from trajectories executed by another agent or robot, although directly transferring the collection of learned skills $\mathcal{A} \backslash \mathcal{A}^0$ is easier and faster.

Learning a new skill amounts to finding a plan $\pi$ achieving a given effect $e$.
Once a successful skill sequence $\{a_0, a_1, .., a_n\}$ was obtained for a given goal $g$, the execution success of each skill $a_i$ can trivially be checked using their effect information. The list of successful skills forms a \emph{trajectory} $\tau$.
Using the initial state $s_0$, $g$ and $\tau$, a new skill can be created with initial plan $\tau$ and intended effect $e$, computed as going from $s_0$ to $g$. Skill conditions $c$ and side effects $e^+$ are computed by composing the conditions and side effects of skills from $\tau$. Learned skills are added to $\mathcal{A}$ and made available to the planner.
Note that this procedure only describes how to initialise a new skill using a successful trajectory. Skills should then be refined to find better plans.

\SetAlgoSkip{MyAlgoSkip}
\begin{algorithm}[t]
 \Fn{plan($s$, $g$, $rec=0$)}{
  \If{$rec > rec_{max} \text{ or } g \text{ statisfied in } s$}{
   \Return{$\text{Random skill }a\text{ from }\mathcal{A}$.}\\
  }
  $a \leftarrow $ skill from $\mathcal{A}$ valid in $s$, with effect closest to $g$.\\
  \eIf{$a \text{ succeeds in state } s$}{
   \Return{$a$.}\\
   }{
   \Return{$\{\text{plan}(s,condition(a),rec+1), a\}$.}\\
   }
 }
 \vspace{0.3em}
 \caption{Goal-regression planner\label{alg:hierarchical_planner}}
\end{algorithm}

\subsection{Skill refinement}
\label{sec:skill_refinement}
After a skill is initialised using a successful trajectory, refining it results in finding a better plan $\pi$ which reuses other high-level skills; see Figure~\ref{fig:graph_skill_refine}.
A given trajectory $\tau$ can be converted to a directed acyclic graph (DAG) $G$ in which nodes are skills and $\tau_0$ is the root. Nodes are linked such that node $n$'s skill condition require all skills in $parents(n)$ to be executed first (ie. the effect of $parents(n)$ are included in node $n$'s skill conditions).
Once graph $G$ is constructed, groups of nodes are replaced by their corresponding higher-level skill from $\mathcal{A} \backslash \mathcal{A}^0$, when their effects and conditions match. After all possible replacements were made, graph $G$ is often composed of few -- and mostly high-level -- skills. Given the trajectory's desired effect $e$, a refined plan $\pi$ can be constructed from $G$ by including all parents of leaf node with effect $e$.
The refinement process is described in Algorithm~\ref{alg:skill_refinement}.

If primitive skill conditions are known, skill refining can be executed directly after initialisation. Conversely, if conditions need to be learned from interaction data, skills may only be refined once the conditions of their successful trajectory sub-skills are confident enough.

\SetAlgoSkip{MyAlgoSkip}
\begin{algorithm}[t]
 \KwData{Trajectory $\tau$, desired effect $e$.}
 \KwResult{Refined plan $\pi$.}
 $G \leftarrow $ build directed acyclic graph from $\tau$.\\
 \For{$\text{node }n \in G$}{
  $a \leftarrow $ shortest skill from $\mathcal{A} \backslash \mathcal{A}^0$ with\\
   \quad effect $\supset n$.effect and condition $\subset n$.condition.\\
  \If{$a \neq \emptyset$}{
   $l \leftarrow $ nodes of $G$ with effect $\subset a$.effect.\\
   Replace nodes $l$ in $G$ with new node from $a$.\\
   }
 }
 $\pi \leftarrow $ skill sequence in $G$ ending with leaf effect $e$.\\
 \vspace{0.4em}
 \caption{Skill refinement\label{alg:skill_refinement}}
\end{algorithm}

\begin{figure*}
\centering
\includegraphics[width=1.6\columnwidth]{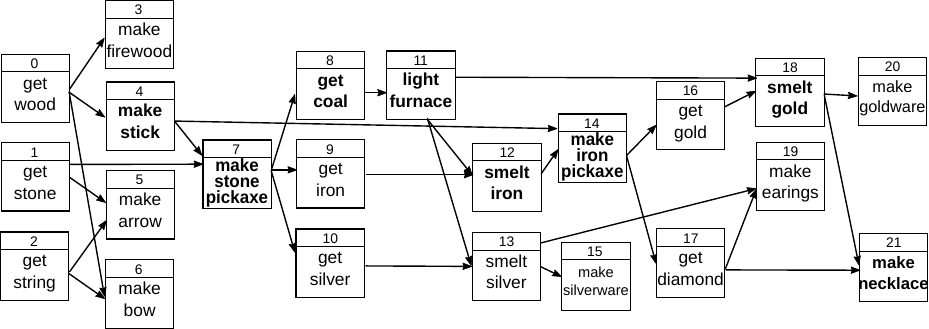}
\caption{Dynamics for Mining domain, and curriculum in bold. Nodes are skills and incoming edges are success conditions.\label{fig:graph_crafting}}
\end{figure*}

\subsection{Condition learning}
\label{sec:condition_learning}
Both planning and skill refinement require skill success condition knowledge to reason about trajectories.
In problems with \emph{unknown} dynamics, the conditions of all primitive skills in $\mathcal{A}^0$ need to be learned. Note that learned skills are composed of lower-level skills and therefore it is not necessary to learn their conditions, as they can be computed by inspecting the skill's plan.

Every interaction with the environment generates a transition tuple $(s, a, s')$. Skill success is easily assessed by comparing the effect $e$ of $a$ to the observed effect between states $s$ and $s'$. After a given primitive skill $a$ is executed $m$ times, starting states and skill successes form a dataset $\mathcal{D}_a = \{s_i, success_i\}_{i=1}^m$ from which conditions can be learned.
Learning a discriminative model $p(success|s)$ of skill success given a starting state is not required to use learned conditions for planning. Rather, learning a generative model $p(s|success=True)$ is more interesting, as it allows the goal-regression planner to use states sampled from the generative model as goals.

Conditions are learned using a two stage process. First, we identify the smallest subset of sufficient dimensions to predict skill success given a state. This is consistent with the framework's requirements of learning sparse skill success conditions.
This step is implemented using orthogonal matching pursuit (OMP) \cite{mallat1993matching,rubinstein2008efficient}.
When applied to $\mathcal{D}_a$, OMP returns coefficients for each dimension of the state space. High coefficients are associated with state dimensions greatly impacting skill success, whereas coefficients close to zero reflect state dimensions with little or no impact. Using the non-zero coefficients returned by OMP, a sparse version of $\mathcal{D}_a$ can be constructed, denoted $\bar{\mathcal{D}}_a$. The second stage of the learning process involves learning a generative model using $\bar{\mathcal{D}}_a$. Any generative model can be used, and we find a Gaussian mixture model (GMM)~\cite{reynolds2015gaussian} is sufficient to learn simple conditions. Increasing the number of components used by GMM allows learning more complicated skill conditions.

\subsection{Learning from curriculum}
\label{sec:curriculum}
Learning good skills is important to reduce planning complexity. Indeed, learning too many skills may result in an unnecessary large search procedure over available skills, whereas learning too few skills leads to flatter planning which requires higher planning horizons.
Deciding when and whether to learn skills is a very challenging problem, and several solutions to address it are proposed in \cite{mcgovern2001automatic,brunskill2014pac}.
While some of these methods could be applied to our work, we chose to focus on learning skills from a curriculum for simplicity. A curriculum is a sequence of goals of increasing complexity, designed by an expert to help learning. The curriculum is composed of a sequence of useful goals, and that mastering earlier goals of the curriculum helps achieving the latter ones. Following this principle, every element of the curriculum is considered a useful skill to learn.
In opposition, learning through random exploration, or even intrinsic exploration \cite{morere2018bayesian}, does not necessarily help discover \emph{useful} goals and skills.

\section{EXPERIMENTS}
\label{sec:experiments}

\begin{table}[t]
	\centering
    \caption{Example of primitive skill conditions learned by the hierarchical agent on the Crafting domain after $41$ training episodes.}
	\begin{tabular}{l | c | c}
	\textbf{Effects} & \textbf{Real conditions} & \textbf{Learned conditions}\\
	\hline
	$s_0 \leftarrow 1$ & None & None\\
	$s_1 \leftarrow 1$ & None & None\\
	$s_2 \leftarrow 1$ & None & None\\
	$s_3 \leftarrow 1$ & $s_0 = 1$ & $s_0 = 1$\\
	$s_4 \leftarrow 1$ & $s_0 = 1$ & $s_0 = 1$\\
	$s_5 \leftarrow 1$ & $s_1 = s_2 = 1$ & $s_1 = s_2 = 1$\\
	$s_6 \leftarrow 1$ & $s_0 = s_2 = 1$ & $s_2 = 1$\\
	$s_7 \leftarrow 1$ & $s_1 = s_4 = 1$ & $s_1 = s_4 = 1$\\
	$s_8 \leftarrow 1$ & $s_7 = 1$ & $s_7 = 1$\\
	$s_9 \leftarrow 1$ & $s_7 = 1$ & $s_7 = 1$\\
	$s_{10} \leftarrow 1$ & $s_7 = 1$ & $s_7 = 1$\\
	$s_{11} \leftarrow 1$ & $s_8 = 1$ & $s_8 = 1$\\
	$s_{12} \leftarrow 1$ & $s_9 = s_{11} = 1$ & $s_9 = s_{11} = 1$\\
	$s_{13} \leftarrow 1$ & $s_{10} = s_{11} = 1$ & $s_{10} = 1$\\
	$s_{14} \leftarrow 1$ & $s_4 = s_{12} = 1$ & $s_{12} = 1$\\
	$s_{15} \leftarrow 1$ & $s_{13} = 1$ & $s_{13} = 1$\\
	$s_{16} \leftarrow 1$ & $s_{14} = 1$ & $s_{14} = 1$\\
	$s_{17} \leftarrow 1$ & $s_{14} = 1$ & $s_{14} = 1$\\
	$s_{18} \leftarrow 1$ & $s_{11} = s_{16}= 1$ & $s_{16} = 1$\\
	$s_{19} \leftarrow 1$ & $s_{13} = s_{17} = 1$ & $s_{17} = 1$\\
	$s_{20} \leftarrow 1$ & $s_{18} = 1$ & $s_{18} = 1$\\
	$s_{21} \leftarrow 1$ & $s_{17} = s_{18} = 1$ & $s_{17} = s_{18} = 1$\\
    \end{tabular}
    \label{tab:learned_conditions}
\end{table}
We present planning results for the presented method in several environments. All environments follow the framework presented in Section~\ref{sec:problem_definition}. As such, dynamics are represented as a DAG, where each node is a primitive skill and connections represent conditions. See Figure~\ref{fig:graph_crafting} for an example.
For simplicity, effects are defined as a state dimension changing from a value of $0$ to $1$.
In all experiments, goals are defined as regions of the state space where some state dimensions are fixed to a desired value. For example, both states $(1,0,1)$ and $(1,0,0)$ match the goal defined as $s_0=1$.

We compare the following methods in experiments.
\emph{MCTS}~\cite{coulom2006efficient} and \emph{RRT}~\cite{lavalle1998rapidly} are non-hierarchical planners that require transition dynamics to be given. They both build a tree of possible futures, by simulating executing skills and their effects on the state. Once a search budget is exhausted, they return the action corresponding to the shortest skill trajectory achieving the goal. We compare MCTS with several search budgets of $100$, $300$ and $1000$, and exploration constant set to $\frac{1}{\sqrt{2}}$. RRT's search mechanism is biased towards the goal with probability $0.1$, and is run for $1000$ and $10000$ steps.

\begin{figure*}
\centering
\includegraphics[width=1.6\columnwidth]{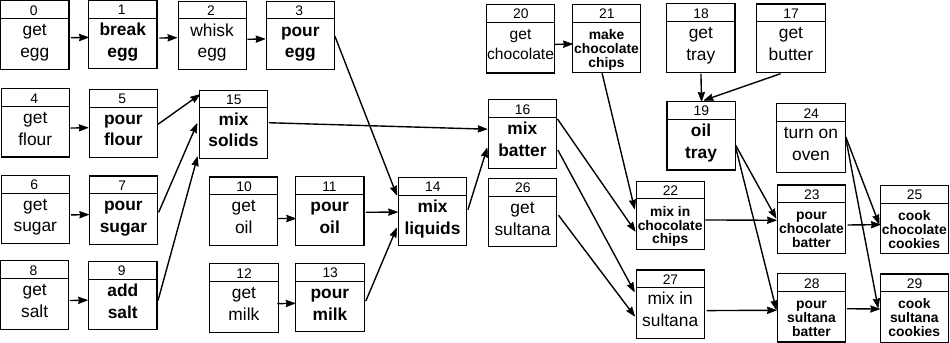}
\caption{Dynamics for Baking domain, and curriculum in bold. Nodes are skills, incoming edges are success conditions.\label{fig:graph_baking}}
\end{figure*}

\emph{Q-learning}~\cite{watkins1992q} is a classic RL method, and does not require the transition dynamics to be specified. A learning procedure is necessary before plans can be generated. Q-learning is trained using a reward function designed for a specific goal (reward of $0$ for reaching the goal and $-1$ otherwise), and thus would need to be trained anew for each goal. In experiments, we allow a maximum of $5000$ learning episodes, which are sequences of up to $100$ actions. If Q-learning successfully finds the goal in $19$ out of the last $20$ episodes, we stop the learning procedure. We use a tabular version of Q-learning, with known convergence guarantees. The discount factor is set to $0.99$, the learning rate to $0.1$, and an epsilon-greedy policy with $0.2$ probability of random action is used.

\emph{DQN}~\cite{mnih2013playing} is a recent extension to Q-learning, using a neural network to model the $Q$ function. The training procedure and parameters are akin to that of Q-learning. One hidden layer with $16$ units is used, and the epsilon-greedy policy parameter linearly decays from $1.0$ to $0.1$.

The \emph{Goal-regression} planner, defined in Algorithm~\ref{alg:hierarchical_planner}, is a simple planner that plans backwards starting from the goal. It recursively plans for the conditions of the last goal until the starting state is reached. In experiments, the maximum recursion of this planner is set to $100$. This planner does not use hierarchical planning and requires known transition dynamics.

Our method, denoted \emph{Hierarchical}, is compared in three configurations. In its simplest form, it requires a list of non-primitive skills to be specified by an expert, as well as transitions to be known; it does not require training. The second variant learns non-primitive skills automatically from data, while still requiring known transition dynamics. The last form learns both skills and transitions from data.
The last two variants both require training with a problem specific curriculum (either a list of goals or demonstrations). As learning skills with known dynamics require a single successful episode, the algorithm advances to the next curriculum stage after its first success on the task. When also learning dynamics, it advances after $5$ successes on a task. The three \emph{Hierarchical} planners are given a maximum planning recursion of $rec_{max}=3$. The GMM used to learn conditions uses $3$ components and the OMP tolerance is set to $3$.

All methods plan and execute one skill at a time, and are terminated if unable to reach the goal after executing $100$ planned skills. Planning results are compared in terms of number of training steps $n_{train}$ (when applicable), plan length $|\pi|$, and time required to generate the plan $t_{plan}$. All algorithms are run on a single CPU core ($2.2$GHz), are averaged over $10$ runs, and results are reported in Table~\ref{tab:results}.

\begin{table}[t]
	\centering
    \caption{Example of skills learned by the hierarchical agent while learning skill conditions (Table~\ref{tab:learned_conditions}) on the Crafting domain after $41$ training episodes. $skill(s_i)$ refers to a learned skill defined in a previous row with effect $s_i \leftarrow 1$.}
	\begin{tabular}{p{1.03cm} | p{1.95cm} | l}
	\textbf{Effect} & \textbf{Plan} & \textbf{Flattened plan}\\
	\hline
	$s_4 \leftarrow 1$ & $0,4$ & $0,4$\\
	$s_7 \leftarrow 1$ & $1, skill(s_4), 7$ & $1, 0, 4, 7$\\
	$s_8 \leftarrow 1$ & $skill(s_7), 8$ & $1, 0, 4, 7, 8$\\
	$s_{11} \leftarrow 1$ & $skill(s_8), 11$ & $1, 0, 4, 7, 8, 11$\\
	$s_{12} \leftarrow 1$ & $skill(s_{11}), 9, 12$ & $1, 0, 4, 7, 8, 11, 9, 12$\\
	$s_{14} \leftarrow 1$ & $skill(s_{12}), 14$ & $1, 0, 4, 7, 8, 11, 9, 12, 14$\\
	$s_{18} \leftarrow 1$ & $skill(s_{14}), 16, 18$ & $1, 0, 4, 7, 8, 11, 9, 12, 14, 16, 18$\\
	$s_{21} \leftarrow 1$ & $skill(s_{18}), 17, 21$ & $1, 0, 4, 7, 8, 11, 9, 12, 14, 16, 18, 17, 21$\\
    \end{tabular}
    \label{tab:learned_skills}
\end{table}

\begin{table*}[t]
	\centering
    \caption{Results in terms of plan length $|\pi|$, number of training episodes $n_{train}$, and planning time $t_{plan}$ in seconds.}
	\begin{tabular}{l c c | c c c | c c c | c c c}
	\textbf{Method} & \textbf{Learning} & \textbf{Learning} & \multicolumn{3}{c|}{\textbf{Crafting} (n=22)} & \multicolumn{3}{c|}{\textbf{Baking} (n=30)} & \multicolumn{3}{c}{\textbf{Random graph} ($n$=100)}\\
	& \textbf{Skills} & \textbf{Transitions} & $n_{train}$ & $|\pi|$ & $t_{plan}$ & $n_{train}$ & $|\pi|$ & $t_{plan}$ & $n_{train}$ & $|\pi|$ & $t_{plan}$\\
	\hline
	MCTS (100) & No & No & -- & 38.9 & 21.3 & -- & 73.0 & 68.1 & -- & 44.0 & 65.3\\
	MCTS (300) & No & No & -- & 24.5 & 34.2 & -- & 55.9 & 135.08 & -- & 85.3 & 318.1\\
	MCTS (1000) & No & No & -- & 27.2 & 115.6 & -- & 48.4 & 362.1 & -- & 75.3 & 8584\\
	RRT (1000) & No & No & -- & 65.4 & 23.1 & -- & 42.4 & 16.9 & -- & 64.8 & 12.7\\
	RRT (10000) & No & No & -- & 45.4 & 123.6 & -- & 29.0 & 86.6 & -- & 45.7 & 87.1\\
	Q-Learning & No & \textbf{Yes} & 580.0 & 70.9 & 0.039 & 5000 & 88.7 & 0.057 & 5000 & 51.7 & 0.129\\
	DQN & No & \textbf{Yes} & 5000 & 94.9 & 0.480 & 5000 & -- & -- & 5000 & -- & --\\
	Goal-regression & No & No & -- & 24.1 & 0.0082 & -- & 39.5 & 0.0183 & -- & 73.67 & 0.063\\
	Hierarchical & No & No & -- & 13 & 0.0384 & -- & 24.0 & 0.104  & -- & -- & --\\
	Hierarchical & \textbf{Yes} & No & 10.0 & 13.1 & 0.0328 & 17.0 & 25.0 & 0.0978 & 17.0 & 25.3 & 0.109\\
	Hierarchical & \textbf{Yes} & \textbf{Yes} & 62.8 & 13.7 & 0.1376 & 96.0 & 29.3 & 0.3714 & 96.1 & 28.7 & 0.340\\
    \end{tabular}
    \label{tab:results}
\end{table*}

\subsection{Crafting problem}
The \emph{Crafting} environment was introduced by \cite{sohn2018hierarchical}, and its deterministic transition dynamics graph is composed of $n=22$ nodes (see Figure~\ref{fig:graph_crafting}) with an average number of $1.3$ conditions per node (see Table~\ref{tab:learned_conditions}). This problem models a robot needing to gather raw materials to craft tools, which are in turn required to gather more advanced materials and craft other objects. Each of the $22$ skills corresponding to graph nodes results in an equivalent state change, eg. executing skill \emph{craft wood stick} changes state predicate \emph{has wood stick} to $True$. This problem features $2^{22} \approx 4.10^6$ different states.

The curriculum used for this problem focuses on graph nodes that are used more often (ie. have more than one child), shown in bold in Figure~\ref{fig:graph_crafting}.
Examples of skills learned using this curriculum with their hierarchical and flattened plans are given in Table~\ref{tab:learned_skills}. The hierarchical plans demonstrate the skill refinement algorithm's capabilities for identifying skills within successful demonstrations and composing learned skills.
Table~\ref{tab:learned_conditions} shows conditions learned by the hierarchical with unknown transition dynamics. While not all conditions are correctly learned, most of them match the real conditions and these are sufficient to generate good plans.

\subsection{Baking problem}
The \emph{Baking} environment models a robot generating plans to bake cookies. Different steps of baking process are included, such as \emph{mix batter} or \emph{oil tray}. The robot is assumed to have previously learned every one of the individual $n=30$ primitive skills of the transition graph shown in Figure~\ref{fig:graph_baking}. From the abstracted skill planning level, the problem is deterministic, and the average number of skill conditions is $1$. The environment has $2^{30} \approx 10^9$ different states.
We generated a curriculum with the same principle as for the previous environment, including skills irrelevant to the testing goal, and omitting some of the steps. The curriculum is shown in bold in Figure~\ref{fig:graph_baking}.

\subsection{Randomly generated problems}
Larger environments are generated by randomly generating directed acyclic task graphs with a fixed number of nodes $n=100$. The number of conditions for each skill is drawn from a Poisson distribution ($\lambda=2$). The generated graphs are denser than that of the previous experiments with an average condition number of $2$, and so less adapted to purely sequential plans. Furthermore, we also introduce stochastic transitions, following Equation \ref{eq:stochastic_transition}, where states are corrupted by switching a random dimension with probability $p=0.2$. This problem has $2^{100} \approx 10^{30}$ states.
The curriculum is automatically generated by ordering non-root transition graph nodes by increasing number of ancestors, only selecting every second node as an intermediary goal.
Because expert skills cannot easily be created for randomly generated transitions, the hierarchical planner with given skills is not run on this problem.

\subsection{Robotics problem: transfer from simulation to real}
\begin{figure}
\centering
\begin{subfigure}{.25\textwidth}
  \centering
  \includegraphics[width=.95\textwidth]{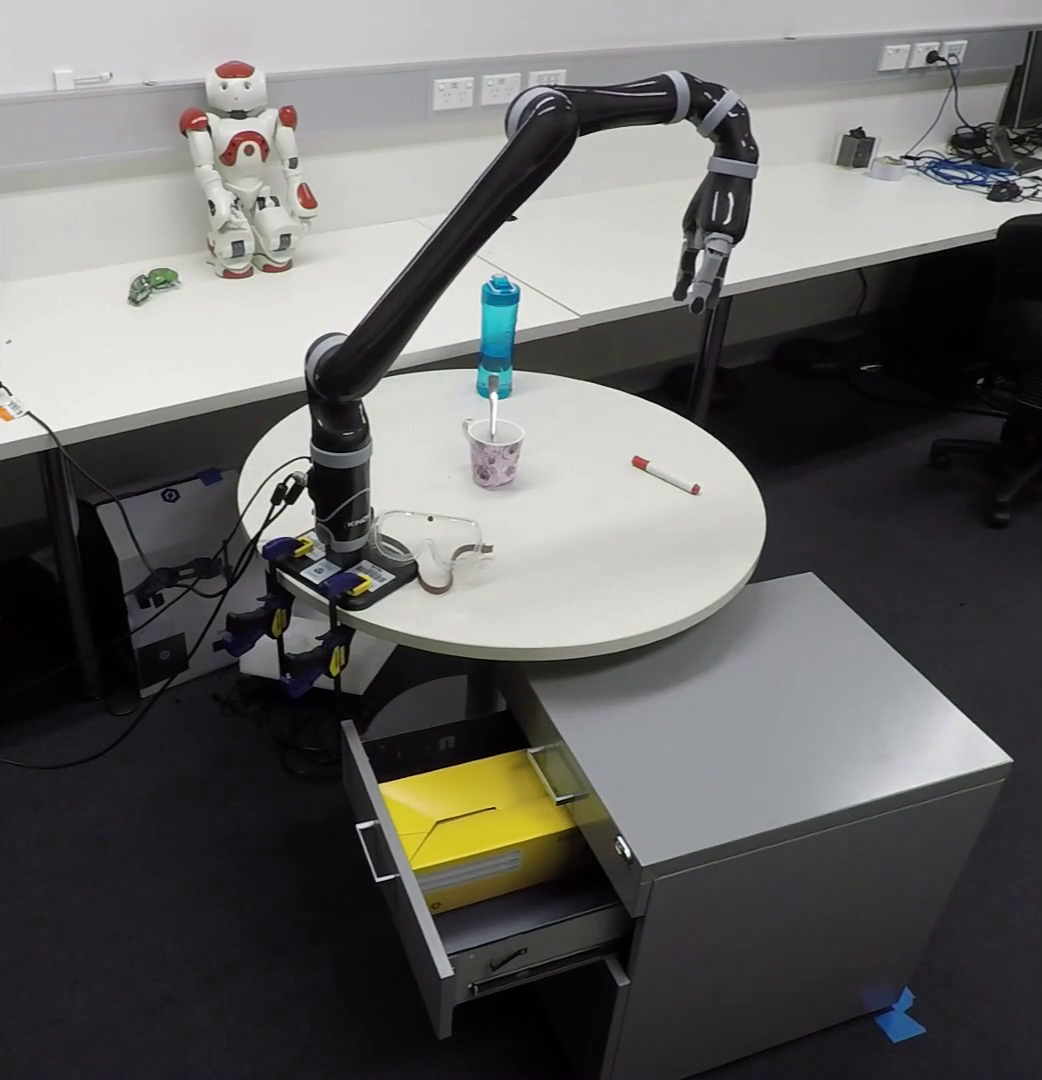}
  \caption{}
  \label{fig:robot}
\end{subfigure}%
\begin{subfigure}{.25\textwidth}
  \centering
  \includegraphics[width=.95\textwidth]{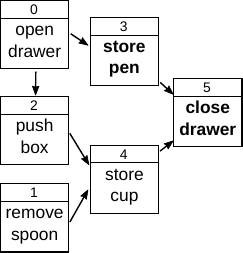}
  \caption{}
  \label{fig:graph_drawer}
\end{subfigure}
\caption{(a) Robotics manipulator used in simulation-to-real Drawer experiment. (b) Dynamics graph for Drawer problem.}
\end{figure}

This last experiment aims to show policies generated in simulation can be applied to a real robot. The problem features a 6 DOF robotics manipulator, shown in Figure~\ref{fig:robot}, aiming to tidy a table by storing items in a drawer. Similarly to previous problems, execution dependencies need to be resolved, eg. a box within the drawer must be pushed to the side before a cup or a pen can be stored next to it. These dependencies are shown in Figure~\ref{fig:graph_drawer}. The manipulator is pre-trained by an expert to execute each individual skill of the graph. This graphical representation is used to learn in simulation, with curriculum $[3,5]$, and generate a \emph{tidy-up} plan. The resulting plan $[0,1,2,4,3,5]$ was generated in $10$ms and is optimal; it is executed on the real robot using pre-trained skills and completes the task, as shown in a supplementary video.

\section{DISCUSSION}
\label{sec:discussion}
Multiple key results can be observed from Table~\ref{tab:results}.
The presented hierarchical planning algorithms (whether learning skills and transitions or not) find shorter plans than the non-hierarchical equivalent \emph{Goal-regression}. This is especially noticeable on problems where the underlying transition graph has increased branching factor like in the \emph{Random} domain. This is because the problems featured in experiments have a hierarchical nature, and not abstracting plans makes planning brittle to small errors when skill effects and conditions are not perfectly matched. 

Planning with goal regression methods is order of magnitude faster than MCTS and RRT. This is to be expected because RRT has no novelty seeking mechanism, and MCTS is not given explicit knowledge of the goal, thus it needs to stumble upon it by chance. This gets worse in problems with higher state dimensions and skill number, whereas goal regression planners don't seem to be affected (as shown by a similar planning time) as they don't need to deal with exploration. For the same reasons, Q-learning and DQN quickly require extensive training time and fail to reach the convergence criteria within the allocated number of training episodes. DQN, which does not have the same convergence guarantees as tabular Q-learning, also failed to find a valid plan on the two more complicated domains.

Learning transitions requires a substantial number of extra learning steps, but ultimately reaches performance similar to the same planner given transitions. Learning skills from curriculum compared to planning with predefined expert skills does not seem to impact planning length much -- provided the given curriculum is good. Learning skills using a curriculum also requires very few learning steps.
Lastly, planning methods seem robust to stochastic transitions, as shown in the \emph{Random} domain. 

Although showing impressive results, the hierarchical planner still has limitations. The environments used in experiments feature relatively simple conditions, defined as the union of a few state dimensions. It would be interesting to extend the condition learning algorithm to handle more complicated conditions.
The current method also requires a curriculum to be designed, conveying what goals are reusable in future tasks. This limitation is relatively mild, given that curricula designed for humans are available for many real-life tasks, which our method could be adapted to leverage.

\addtolength{\textheight}{-2.4cm} 

\section{CONCLUSION AND FUTURE WORK}
\label{sec:conclusion}
We presented a framework and algorithm for learning to plan hierarchically in domains with unknown dynamics, in which transition dynamics are decomposed into skill effects and conditions. We proposed a method for learning both abstract skills and their success conditions from interaction with the environment.
We validated our method in experiments of increasing complexity (with up to $2^{100}$ states), demonstrating superior planning to classic non-hierarchical planners or reinforcement learning methods. Lastly, we showed the algorithm is applicable to robotics problems in a simulation-to-real transfer problem.

The presented algorithm is also able to interact with humans. Because skill conditions are explicitly modelled, the planner is aware of what it doesn't know and can ask queries such as \emph{What do I need to craft a pickaxe?}. Human responses as conditions (\emph{get a wood stick and stone}) or a sequence of goals (\emph{craft a wood stick, get stone, craft pickaxe}) can both be used to refine skill conditions.

The avenues for future work are numerous:
Learning conditions using causal learning \cite{holland1986statistics} or inductive logic programming \cite{muggleton1994inductive} could be greatly beneficial. Having a causal model would allow agents to actively explore the space of possible transition dynamics, by choosing actions that run causal tests. This kind of active exploration could result in significant improvement over choosing random actions.
Generating and updating a curriculum automatically would reduce the amount of expert knowledge required to apply our method. The work of \cite{florensa2018automatic} uses generative adversarial networks to automatically generate challenging goals, and could be combined with our work.
It is yet unclear how the presented algorithm would scale to much larger problems such as life-long learning. Skill forgetting and/or better refinement would probably be necessary to tackle these challenging problems.
Although our framework features sparse state spaces, more domain structure could be exploited using concepts from the oriented-object MDP framework~\cite{diuk2008object}. This would enable reasoning about object function, properties and skill affordances~\cite{sridharan2017can}. Actions available to the agent could also be restricted depending on the objects observed in the state space \cite{nair2018action}.




\bibliographystyle{plain}
\bibliography{references}
\end{document}